# SIND: A Drone Dataset at Signalized Intersection in China

Yanchao Xu, Wenbo Shao, Jun Li, Kai Yang, Weida Wang, Hua Huang, Chen Lv, Hong Wang*

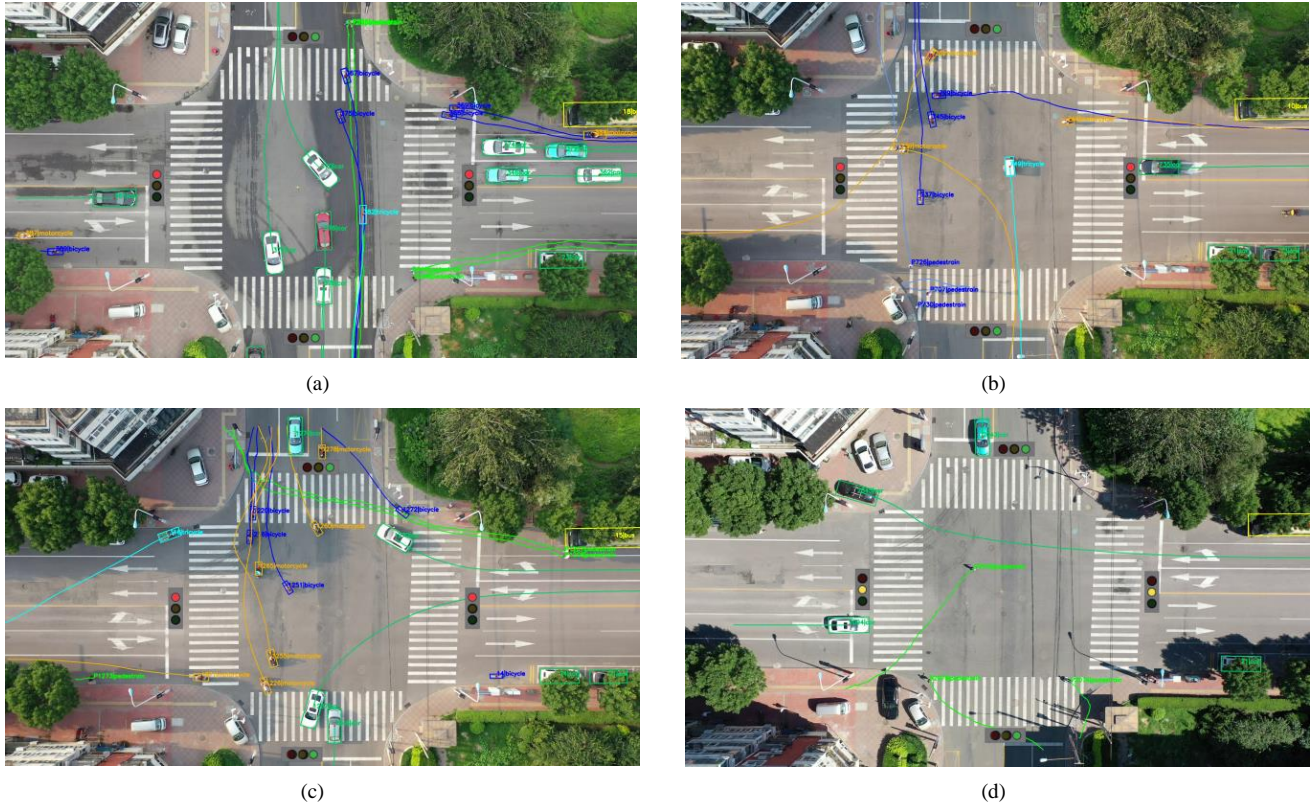

Fig. 1: A few exemplary visualizations of the SIND dataset. For each traffic participant in the frame, the previous trajectory is plotted. There are some typical scenarios: (a) Conflict between unprotected left-turn vehicles and straight-going vehicles (b) Non-motor vehicle running the red light (c) Multi-non-motor vehicle mixed traffic scenarios (d) Motor vehicle runs a yellow light and special behavior(Pedestrians cross the intersection diagonally, vehicles drive onto the sidewalk)

*Abstract*—Intersection is one of the most challenging scenarios for autonomous driving tasks. Due to the complexity and stochasticity, essential applications (e.g., behavior modeling, motion prediction, safety validation, etc.) at intersections rely heavily on data-driven techniques. Thus, there is an intense demand for trajectory datasets of traffic participants (TPs) in intersections. Currently, most intersections in urban areas are equipped with traffic lights. However, there is not yet a large-scale, high-quality, publicly available trajectory dataset for signalized intersections. Therefore, in this paper, a typical two-phase signalized intersection is selected in Tianjin, China. Besides, a pipeline is designed to construct a Signalized INtersection Dataset (SIND), which contains 7 hours of recording including over 13,000 TPs with 7 types. Then, the behaviors of traffic light violations in SIND are recorded. Furthermore, the SIND is also compared with other similar works. The features of the SIND can be summarized as follows:

1) SIND provides more comprehensive information, including traffic light states, motion parameters, High Definition (HD) map, etc. 2) The category of TPs is diverse and characteristic, where the proportion of vulnerable road users (VRUs) is up to 62.6% 3) Multiple traffic light violations of non-motor vehicles are shown.

We believe that SIND would be an effective supplement to existing datasets and can promote related research on autonomous driving.

The dataset is available online via:
https://github.com/SOTIF-AVLab/SinD

* This work was supported by the National Key R&D Program of China:2020YFB1600303, the National Science Foundation of China Project: U1964203 and 52072215.

Yanchao Xu, Weida Wang, Hua Huang are from the School of Mechanical Engineering, Beijing Institute of Technology, Beijing, China. (E-mails: 3120200410@bit.edu.cn, wangwd0430@bit.edu.cn, 13910848350@139.com ).

Hong Wang, Wenbo Shao, Jun Li are with the School of Vehicle and Mobility, Tsinghua University, Beijing, China. (E-mails: hong_wang@tsinghua.edu.cn, swb19@mails.tsinghua.edu.cn, lijun1958@tsinghua.edu.cn)

Chen Lv is with the Nanyang Technological University, Singa (E-mail: lyuchen@ntu.edu.sg ).

Kai Yang is with College of Mechanical and Vehicle Engineering, Chongqing University, Chongqing, China (E-mail: kaiyang0401@gmail.com )

(Corresponding author: Hong Wang)

## I. INTRODUCTION

Autonomous driving is considered one of the revolutionary technologies shaping humanity's future mobility and quality of life. However, safety remains a critical challenge for the commercialization and widespread application of autonomous vehicles. A key challenge to ensure the safety of autonomous vehicles is to comprehensively understand the complex traffic context and accurately predict the motion of the traffic participants (TPs). Presently, the intersection is a critical scenario for autonomous vehicles due to the sophisticated topology structures and dynamic traffic conditions [1].

The current motion prediction solutions are designed with a strong reliance on data-driven methods (e.g., deep learning techniques) where a large amount of realistic traffic data are required. Over the past decades, multiple studies assume that the future motion of TPs is mainly determined by the historical trajectories[2-3]. Actually, infrastructure (e.g., road structure), traffic rules ( e.g., zebra crossings, stop lines, and right of way) will also significantly impact the motion of TPs. For these reasons, the state-of-the-art methods utilize raster graphics or vector maps to encode the road infrastructure to motion prediction model[4-7]. In particular, at the signalized intersections, the states of traffic lights directly affect the motions of TPs. Therefore, it is crucial to integrate the states of traffic lights into the motion prediction model to improve the performance as done in[8-10]. On the other hand, autonomous vehicles ought to behave similarly to human-driven vehicles to be more predictable to others. By investigating the interaction mechanism of TPs from the realistic traffic dataset, imitation learning or human-like decision-making can be achieved[13-14]. In other words, comprehensive traffic patterns, e.g., road maps and traffic lights can be utilized to facilitate safe and socially compliant behavior [12].

To address the aforementioned issues, interactive trajectory datasets recorded from real-world driving scenarios are urgently required, especially those consisting of traffic lights, road maps, etc. Therefore, in this paper, trajectory data of a signalized intersection in Tianjin, China will be collected via drone to fill the gap that most existing datasets lack traffic lights information. As explained in [15] [17], a drone hovering at a high height will have little effect on TPs on the ground, allowing the high-definition camera carried by the drone to capture their naturalistic behaviors. Specifically, SIND, as shown in Fig.1, consists of two-phase traffic lights, a large number of Vulnerable Road Users (VRUs), and some special behaviors of TPs. Moreover, the trajectories of TPs and the High Definition (HD) map are also available in the SIND dataset. Furthermore, the behavior types of vehicles (turn left, turn right, go straight, and others) are labeled according to the HD map and traffic lights states and the frequency of traffic light violations is also shown in SIND. Moreover, SIND is also compared with two similar datasets, which distinguishes SIND from others. SIND has promising applications in many research fields including motion prediction in mixed traffic [18], behavior modeling [19], scenario construction [20], test case generation[21], traffic simulation [23], etc. Finally, multiple critical behaviors in SIND as shown in Fig. 1(d) are of great importance for the research on autonomous driving, e.g., responsibility-sensitive safety (RSS) [24], Safety-considering decision-making[25] and trajectory planning [26][27], and ethical decision-making[28-29].

## II. RELATED WORK

In recent years, there are many publicly available trajectory datasets to facilitate the research on autonomous driving. These datasets are constructed via different approaches in diverse locations. The well-known Next Generation Simulation (NGSIM) [30] is the first publicly available large-scale trajectory dataset, which was recorded via cameras mounted on the top of buildings. Specifically, NGSIM was recorded at four different locations, including a two-way urban arterial with four signalized intersections at the Lankershim Boulevard (subsequently referred to LB) in Los Angeles, CA. However, the LB dataset only comprises the trajectories of vehicles, but the motion of VRUs is not captured, which is the difference compared with SIND. Besides, the recording time of LB is only 30 minutes. Overall, the NGSIM dataset can not meet the needs of the current research on autonomous driving tasks well, which is mainly oriented to traffic simulation [31].

To facilitate the research on the urban traffic environment, the INTERACTION Dataset was collected, which is the first large-scale drone dataset for the TPs in the urban areas [15]. Concretely, The INTERACTION dataset contains the naturalistic motions of various traffic participants. It contains a variety of highly interactive driving scenarios from different countries, i.e., China, Germany, and the United States. This dataset covers four types of road structures, i.e., roundabouts, unsignalized intersections, signalized intersections, ramp merging & lane changes. The recording time is up to 16.5 hours, which is the first to provide HD-map information in lanelet2 format [16]. Nonetheless, there are only a few categories of TPs in INTERACTION. In particular, the proportion of VRUs is low (only about 5%), and the motion features of TPs in INTERACTION are relatively insufficient. For instance, only the velocity in global coordinates is provided. More importantly, this dataset does not provide any information on the traffic lights at signalized intersections, which is not conducive for researchers to conduct studies related to signalized intersections.

Analogously, another urban dataset, i.e., the inD dataset has been published, which was recorded at four different unsignalized intersections in Aachen, Germany. It contains a total of 13,599 trajectories over 10 hours [17]. The inD dataset classifies the TPs into four types, i.e., car, truck/bus, pedestrian, and bicycle, which has a higher proportion of VRUs (about 40%) than INTERACTION. Apart from the features of position, heading angle, and object size (pedestrians and bicycles have no size), the features of longitudinal and lateral acceleration are also provided in the inD dataset. Besides, it also contains metadata about the recording and TPs, such as recording time, sites, etc. In addition, there are also various roundabout datasets, e.g., rounD [32] and openDD [33] where rounD includes 13746 trajectories, and OpenDD contains 84774 trajectories. Nevertheless, both rounD and openDD are recorded in the unsignalized roundabout and the proportion of VRUs is less than 5%.

On the other hand, several large-scale trajectory datasets have been collected via onboard-sensor-equipped fleets[34-

37]. These datasets only contain separated trajectory segments of the intersection scenarios，which limits the application of datasets in traffic simulation and other fields. Furthermore, completeness is also another key issue for this type of dataset. Overall, onboard-sensor-based datasets cannot fully replace the role of drone datasets.

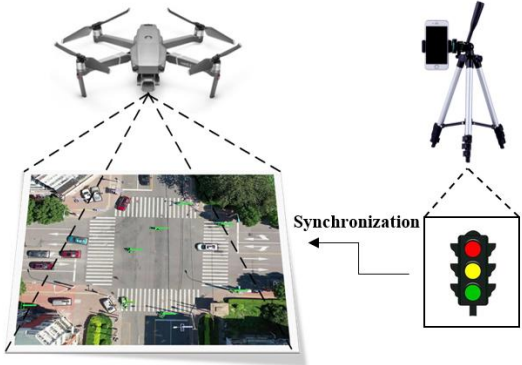

Fig. 2: Drones were used with HD cameras to record traffic and use algorithms to generate trajectories. Another camera placed on the ground was used to record traffic light states. Synchronization between trajectories and traffic light states is required.

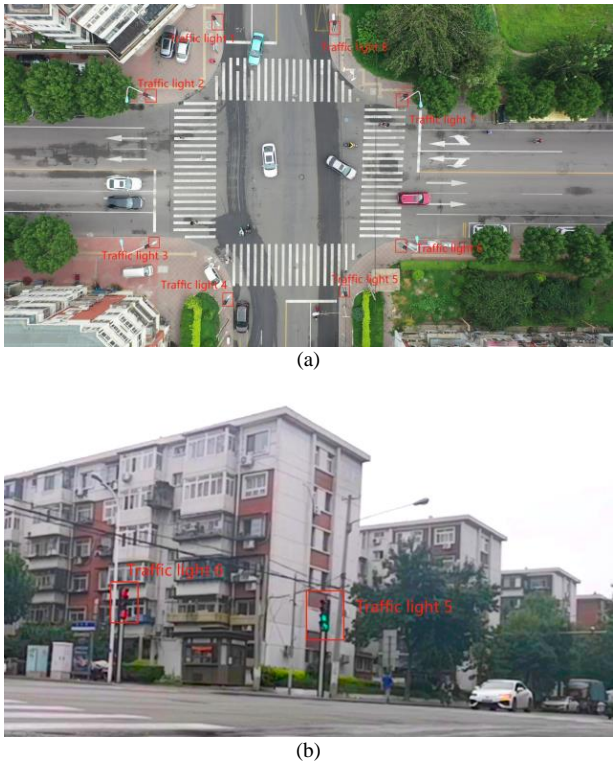

Fig. 3: Selected intersection and corresponding traffic lights. (a) The location of the traffic light at the intersection (b)Type and appearance of traffic lights

## III. CONSTRUCTION OF DATASET

In this section, the pipeline used for dataset construction is shown, which covers location selection, data recording, data processing, maps generation, etc.

### A. Selection of recording location

From the perspective of functional requirements, the recorded scenario should be generic and challenging for autonomous driving tasks. On the one hand, the intersection structure and traffic light type should be general with normal traffic patterns. On the other hand, it is expected that there are some non-compliant behaviors of TPs, which are challenging for motion prediction and decision-making tasks. In addition, it is essential to note that the intersection cannot be located in a no-fly zone. Thus, a signalized intersection with two-phase traffic lights is selected, which is located in Tianjin, China, as shown in Fig. 3. In detail, this intersection is controlled by two-phase traffic lights, where turning left and going straight share the same green light phase, which leads to frequent interactions and conflicts between TPs. In this context, a higher number of VRUs, including pedestrians and non-motor vehicles with their violations, are recorded in this dataset.

### B. Data recording and preprocessing

Similar to previous works, the trajectory data in this intersection is also recorded via drone. Specifically, a DJI Mavic 2 equipped with a 4K (3840x2160) camera is utilized that captures trajectories at 29.97 Hz. Another significant contribution of this dataset is that the traffic light states are provided. Actually, the state of the traffic light can not be extracted from the drone's views. Therefore, another camera was posed on the corners of the sidewalk to record the variation of the traffic light states. The steps for data collection and preprocessing are listed as follows.

- **Data recording**: The drone hovers at a height of 75m to record the intersection. The hover height is carefully chosen for view coverage and image clarity. The weather condition is also considered when recording data. These recordings cover the morning and afternoon periods, as well as sunny days, cloudy days, and after rain. A total of 23 recordings are made in the span of 15 days, ranging from 10 to 22 minutes.

- **Time synchronization**: Each captured video has a synchronization frame when the corresponding camera recording the traffic light states. The video editing software can be used to align the time axis and set a unified time stamp.

- **Downsampling and stabilization**: To tackle the issue that TPs move slowly between frames in the original video, the downsampling technique is applied at equal frame intervals and one-third of the frames are kept as images. In order to eliminate the disturbance of slight movements of the drone, a matching algorithm based on the Speeded Up Robust Features (SURF) [38] is used, which projects each frame to the first frame of the record, and further registers it with the HD map.

### C. Detection and correction

- **Detection**: To create trajectories, the class and position of all TPs in each frame should be obtained.

Besides, size and orientation are important features for vehicles. In SIND, pedestrians are annotated as points, and vehicles are presented as axis-oriented rotated rectangles on each frame. To improve the efficiency of annotations, the YOLOv5 [39]-based method is utilized to identify objects and generate rotated rectangles. In this process, there are only a few cases (less than 5%) where detection is inaccurate, which can be resolved by manual correction.

- **Relief displacement correction**: Owing to the hovering height of the drone is relatively low, the imaging process cannot be regarded as an orthographic projection. Because the height of the object will also cause its outward offset from the imaging center [40]. This is especially evident for trucks and buses with high-body. Thus, the method in [41] is applied to eliminate its impacts.

*D. Tracking and Postprocessing*

After annotating rotated rectangles in each frame, it is necessary to assign uniform IDs to bounding boxes of the TPs, namely tracking. Therefore, the TPs' position in the pixel coordinate system should be transformed to the ground coordinate system. It also should be aligned with the HD map to represent the ground infrastructure. Finally, the model-based smoothing method is used to refine the trajectory and estimate parameters e.g., velocity and acceleration.

- **Tracking**: The method based on intersection-over-union (IOU) matching and linear Kalman filtering is used to track [42]. Due to occlusions such as tree shade, several tracks of TPs are missing. To tackle this issue, the missing positions are obtained via prediction techniques. For a few failure cases, it is effective to manually check whether there are multiple tracks belonging to the same TPs, merging them if true.

- **Coordinate Systems and HD Maps**: The ground coordinate system is utilized, and the marked point on the ground is taken as the origin. A frame is selected as the reference image, and based on the reference image, the HD-map is drawn according to lanelet2 format. Note that all TPs and map elements should be converted to the ground coordinate system.

- **Coordinate conversion**: Due to the stabilization of the drone gimbal, it is assumed that the effect of the camera pitch angle could be ignored, and tracks in all records can be converted to the ground coordinate system by a similar transformation. Scale is calculated by measuring ground reference elements, and translation and rotation parameters can be obtained by registering all recorded first frames with the benchmark.

- **Smoothing**: Rauch-Tung-Striebel(RTS) smoothing [43] method is used to refine the trajectory by considering kinematic constraints. Here, the bicycle model [44] and the point mass model [45] are used for vehicles and pedestrians, respectively. The influence of detection error is reduced by the smoothing technique, and the motion states e.g., velocity and acceleration are estimated simultaneously.

*E. Datasets and Formats[1]*

With the above steps, the SIND dataset can be constructed. The following characteristics distinguish our work from others.

- **Trajectories and motion state:** The motion features of TPs including position, velocity, and acceleration are recorded in CSV files. The details of pedestrians and vehicles are stored separately. For vehicles, the motion states in the ego-vehicle coordinate system, e.g., heading angle and the yaw angle are also included.

- **Traffic light states:** SIND includes 8 traffic lights and corresponding states. Specifically, 0 means "red", 1 means "green" and 3 means "yellow".

- **HD Map:** A lanelet2 map with the OSM format and an aerial image are used to describe static elements of the selected intersection, as demonstrated in Fig. 4.

- **Tracks Meta-information:** The start frame and end frame of TPs in each record are saved in the meta CSV file. Besides, crossing types of vehicles are classified and the traffic light violations are also marked (see section IV). In addition, the specific collection time, weather conditions, the number of traffic participants are also given in the recording meta CSV file.

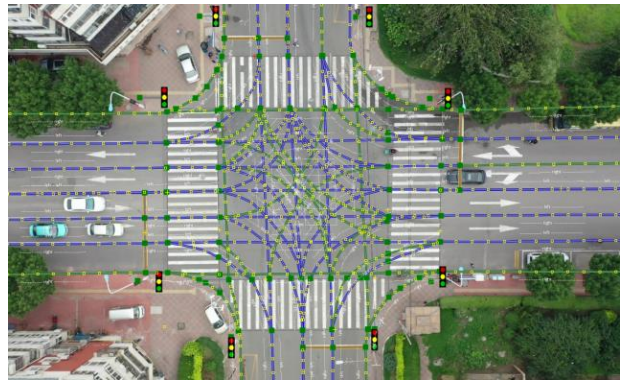

Fig. 4: Lanelet2 format map of the signalized intersection.

## IV. STATISTICS AND EVALUATION

*A. Scale and number of categories*

The SIND contains 23 records, each with a duration of 8-20 minutes, for a total of 420 minutes. The recording time covers the morning and afternoon between July 28, 2021 and August 11, 2021. The weather during that time was sunny, cloudy, or after rain. SIND dataset contains 13,248 trajectories in total, recording more than 99.8% of the TPs at the intersection (except for an extremely small number of special objects such as pets and carts). Classes of TPs include car, truck, bus, tricycle, bike, motorcycle, and pedestrians, as shown in Fig. 5.

---

[1] For the detailed format of SIND, see https://github.com/SOTIF-AVLab/SinD/blob/main/Format.md

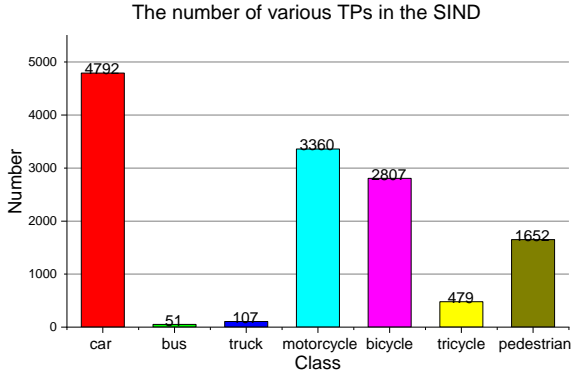

Fig. 5: Number of TPs in each class in SIND. The motorcycle, bicycle, tricycle, and pedestrian classes belong to the VRUs.

*B. Traffic light violations*

Traffic light violations are significant causes of traffic accidents, and research on these violations can undoubtedly benefit safe autonomous driving. Therefore, traffic light violations of vehicles and non-motor vehicles at the selected intersection have been identified. According to the regulations of China's Road Traffic Law [46], traffic light violations by motor vehicles can be defined as:

- **Red light violation**: When the front wheels of the vehicle cross the stop line, the red light comes on and the rear wheels continue to cross the stop line and the vehicle drives directly to the opposite side (the front wheels crosse the stop line on the opposite side). Note that it belongs to over-the-line parking if the vehicle doesn't drive to the opposite side, which is very rare, so it will not be discussed separately.

- **Yellow light violation**: When the front wheels crosse the stop line, the yellow light is already on, and then the vehicle drives directly to the opposite side.

Firstly, the HD map is utilized to judge the behavior of vehicles including turning left, turning right, going straight, or other behaviors, e.g., driving to the sidewalk. Then, the violation is be identified based on the traffic light states and the frame time when the vehicle crosses the stop line. Here, only vehicles turning left and going straight need to be considered because vehicles turning right are allowed all the time in China. Besides, the rules for non-motor vehicles are similar, except that the stop line is replaced by the border of the zebra crossing because non-motor vehicles can stop in this area. The number of traffic light violations has been counted in the whole dataset as shown in Fig. 6. For each record, vehicle crossing types and traffic light violations are also provided in the tracks meta files.

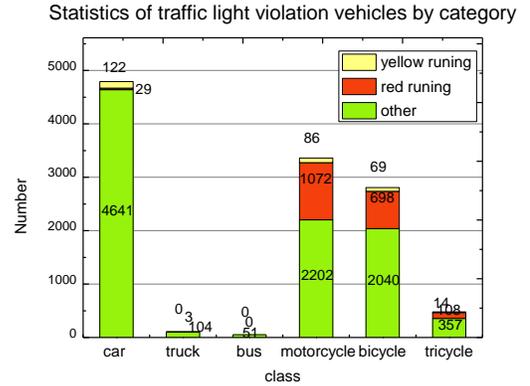

Fig. 6: Statistics of vehicles violating traffic light signals in SIND. Red represents the behavior of running red lights, yellow represents the behavior of running yellow lights, and green represents other behaviors that do not violate traffic lights. Count of violations by category: car(29/122), bus(0/0), truck(3/0), biycle(698/69), motorcyle(1072/86), tricycle(108/14).

*C. Comparison and Discussion*

As aforementioned, although there are several drone datasets, this paper focuses on a signalized intersection with diverse TPs. Therefore, to show the difference, the SIND dataset is compared with two of the recent typical homogeneous datasets. The comparative results are shown in Table I and Fig. 7.

It can be found that all three datasets provide the HD map, which is significant for analyzing and predicting the behavior of TPs. However, these datasets are quite different in other aspects. Firstly, INTERACTION dataset includes diverse sites and is also with the largest scale. It contains 18,471 trajectories of TPs at intersections, where 14,867 trajectories are recorded at unsignaled intersections and 3,775 trajectories belong to signalized intersections. However, there is no signal state information in the whole INTERACTION data. Besides, it is unreasonable that the signalized intersections should contain a large number of pedestrians according to the provided snapshot. But the corresponding files do not provide any information about the pedestrian/bicycle, which makes the completeness of the datasets questionable.

In addition, the inD dataset was collected from four unsignalized intersections in Aachen, Germany. Compared with INTERACTION, inD has more types of TPs, especially existing works. In SIND, the heading and yaw angles of the vehicle are also distinguished as follows.

TABLE I: Comparison with the urban intersection dataset.

| Dataset name | Length | Location | Collection Site | Trajectories | Trajectory types | HD Map | Traffic light | Sample Freq |
|---|---|---|---|---|---|---|---|---|
| INTERACTION | 16.50 hours | Urban intersections, Roundabout, Highway | 11 | 40054 | Pedestrian/bicycle, and car | lanelet2 | no | 10Hz |
| InD | 10.00 hours | Urban intersections | 4 | 13599 | pedestrian, bicycle, car, and bus | lanelet2 | no | 25Hz |
| SIND | 7.02 hours | Urban intersections | 1 | about 13248 | Car,bus,truck,bicycle, motorcycle,tricycle | lanelet2 | yes | 10Hz |

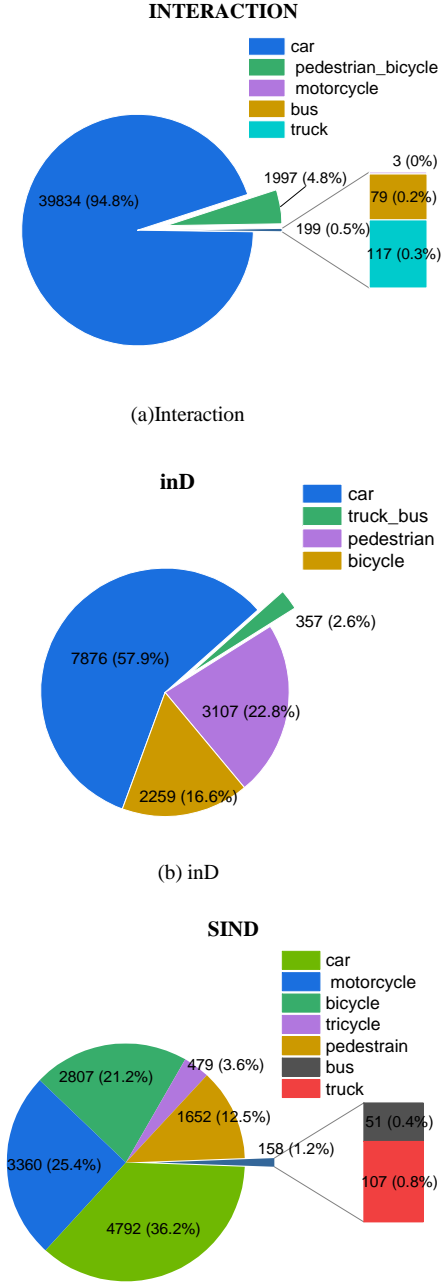

Fig. 7: Comparison of the number and proportion of categories: INTERACTION, inD, SinD.

$$\theta = \beta + \psi \qquad (1)$$

where $\theta$ is the heading angle, $\psi$ is the yaw angle, $\beta$ is the sideslip angle of the center of mass. The heading angle of vehicles refers to the angle between the vehicle centroid speed and the horizontal axis in the ground coordinate system. Yaw angle refers to the angle between the vehicle's longitudinal axis and the horizontal axis.

For both motor and non-motor vehicles, the comprehensive size and motion state features are given. Specifically, there are 7 types of traffic participants in SIND and a particularly high proportion of VRUs (e.g., the trajectories of pedestrians, bicycles, motorcycles, and tricycles account for about 62.6%, which is much higher than that of INTERACTION and inD.)

For the quality comparison, there are no results shown in INTERACTION to verify the accuracy. inD provides a short demonstration video on its website[2]. From a glance at the video, it is accurate, but the little jump in the vehicle's orientation can still be observed. Accordingly, a corresponding demo video of the SIND is also available online[3]. From the visual observation of the video recording, it can be concluded that the quality of SIND is close to inD, or even better than it, especially for small-sized bicycles and motorcycles.

As aforementioned, SIND is collected at signalized intersections and provides comprehensive information. At signalized intersections, both traffic light status and high-definition maps are significant features for motion prediction, behavioral decision-making, and scenario modeling tasks. In addition, the unprotected left turns are also considered one of the challenging scenarios for autonomous driving, which also requires a large amount of real-world data. At a two-phase signalized intersection, left-turning and straight-going vehicles share one signal phase, and left-turning vehicles have no right of way for straight-driving vehicles. Correspondingly, there are many interactions and conflicts between left-turning and straight-going vehicles at the two-phase signalized intersection. Obviously, SIND naturally contains many similar situations. Meanwhile, other related research demands can be also well satisfied by SIND.

More importantly, the vehicle traffic light violations are recorded in SIND. To the best of our knowledge, there is currently no similar dataset for statistical comparison. A large number of traffic light violations of non-motor vehicles at this intersection are provided, the proportion of which far exceeds that of motor vehicles. Violations are strongly related to the size of the intersection and the flow of traffic. In general, smaller intersections and sparse traffic are more likely to lead to illegal crossing behaviors.

VRUs. And it provides more motion state features including acceleration, which is not only recorded in the ground coordinate system but also in each ego-vehicle coordinate system. The sampling frequency of inD is up to 25Hz and thus, the motion states estimation can be more accurate. However, inD does not provide the information of size and orientation for the bicycle class.

Overall, the scale of SIND is close to that of inD, but SIND has more intensive traffic activities. More importantly, SIND has the most complete motion parameters compared with

## V. CONCLUSION

In this paper, a drone dataset SIND at the signalized intersection in Tianjin, China was collected. Firstly, a pipeline was designed to construct a trajectory dataset at the selected intersection. With the pipeline, SIND was constructed, which provides comprehensive motion features of TPs, including traffic light states, and HD maps. More importantly, the traffic light violation behaviors of vehicles in SIND were labeled according to the traffic rules in China, which distinguishes SIND from other works. Furthermore, SIND includes a high proportion of VRUs, and the traffic light violations of non-motor vehicles are also given. Finally, we believe that SIND will facilitate the research on autonomous driving, especially motion prediction in mixed traffic, behavioral modeling of VRUs, etc. The full dataset will be released before the conference.

---

[2] https://www.ind-dataset.com/

[3] SIND demo video

## VI. LIMITATION AND FUTURE WORK

The current data is collected in a single location, which is not conducive to analysing the impact of location and road structure on the behavior of TPs. Therefore, in the next phase of work, we plan to collect data on different intersections in various cities of China considering factors such as regions characteristics, intersection types, to construct larger-scale and diverse urban trajectory dataset, covering more scenarios of interest in autonoumous driving.

In addition, relative to vehicles, the trajectories of pedestrians are difficult to simply generalize by several types due to inherent complexity, so we currently do not provide labels for their behavior types and violation types. The generalization and labelling of pedestrian behavior at intersections will also be part of the follow-up work.

While striving to collect and build large-scale and diverse data, we will also advance the corresponding urban autonomous driving scenario tests, behavioral analysis of VRUs, and research on prediction and decision-making algorithms in comprehensive scenarios. These results will be released in subsequent publications.